# Finding Emotions in Faces: A Meta-Classifier

Siddartha Dalal
Columbia University
New York, NY

Michael Lesk
Rutgers University
New Brunswick, NJ

Sierra Vo
Columbia University
New York, NY

Wesley Yuan
Columbia University
New York, NY

*Abstract*—Machine learning has been used to recognize emotions in faces, typically by looking for 8 different emotional states (neutral, happy, sad, surprise, fear, disgust, anger and contempt). We consider two approaches: feature recognition based on facial landmarks and deep learning on all pixels; each produced 58% overall accuracy. However, they produced different results on different images and thus we propose a new meta-classifier combining these approaches. It produces far better results with 77% accuracy.

Keywords— emotion recognition, facial landmarks, combined learning methods.

## I. INTRODUCTION

Earlier researchers have tried to project face emotions in 8 categories proposed by Ekman (1993): neutral, happy, sad, surprise, fear, disgust, anger and contempt. Besides human raters two machine learning methods have been used to detect the emotions in these 8 categories: (a) by using a small number of features using facial landmarks (Martinez 2016), and b) deep learning based methods using all the pixels of the face (Mellouk and Handouzi 2020). There are many applications, such as medicine (Harms 2010) and recognizing fatigue (Jeong 2018).

Despite much prior work on both static pictures and on video data, it is unclear which facial characteristics should be singled out as important. Even if we knew which features to use, there are multiple algorithms that might be used for learning. Previous research (Cheng 2016, and Lian 2018) has looked at combining different methods of machine learning. In this paper we study the choice between landmark-based features of a face and performing deep learning on full images.

For face emotion detection, there are several databases where the ground truth is human rated (Naga 2021). The largest database of this type is Affectnet. Affectnet is a dataset of about a million faces (swept from the web) of which more than 400,000 had emotions assigned by twelve human annotators (Molehassani 2019). Each image is a single face, unobstructed, and cropped to remove extraneous background. The emotional categories are the standard ones defined by Ekman, and the judgments are forced-choice (only one emotion for each face). In our work we only used the manually assigned emotions; the full Affectnet also has hundreds of thousands of automatically categorized pictures.

The emotions in Affectnet are unevenly distributed, with more than 140,000 faces labeled as "happy" but only about 5,000 as either "contempt" or "disgust". We selected a balanced subset with the same number of images for each emotion, split it 80-20 randomly into a training and test set, and then deleted duplicates, ending with 16,854 training images and 4,520 test images. We checked for bias in the affectnet data and found no relationship between image face color and judgment of valence.

We began by employing and comparing a landmark-based approach and a deep learning pixel-based approach. We show how both methods can be complementary and can be combined to improve the classification.

## II. PREPROCESSING FOR ALIGNING FACES USING LANDMARK-BASED APPROACH

For landmark analysis it is necessary to align faces. There are a number of different techniques used for these including using a 3d model of human faces and aligning a given image to match in shape with the 3d model Zhu, Liu, Lei and Li (2019) and Su, Ai and Lao (2009). An advantage of that approach is that one can find the occluded landmarks. However, the methods are rather complex and it is not clear that the solutions are robust in finding the exact occluded areas and the corresponding emotions.

Instead our objective is mainly to identify emotions in faces which are in portrait modes. However, the faces need to be aligned by translating, scaling, rotating and resizing before applying classifiers.

Our algorithm for face alignment consisted of:

1) extract the faces by using bounding boxes (MTCNN- deep learning algorithms),

2) resize the image under consideration to be 200*200,

3) use DLIB landmark detection algorithm to identify landmarks coordinates. These gives coordinates of various features including tip and bridge of the nose, eyes, etc.

4) move the landmarks so that the tip of the nose is at the center of the image. This centers the image.

5) However, the faces may be leaning. Thus, find the angle between the line joining the tip of the nose to the bridge of the nose to the perpendicular line

6) rotate the angle so that line now matches the perpendicular line (See the attached Figure 1- landmark 27 is the bridge and landmark 33 is tip of the nose).

7) finally standardize the length of the nose and the distance between the left and right corner of the left eye are fixed.

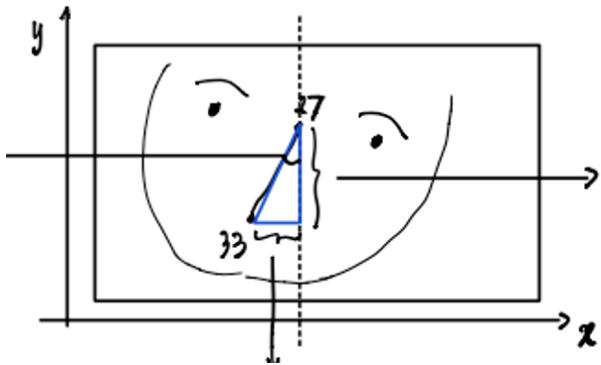

Figure 1: Rotation to straighten out face

### III. LANDMARK BASED RESULTS

Starting with the standardized images, we chose features by computing distances and angles between landmark points. The AffectNet images come with 68 landmarks identified using OpenCV. See Figure 2.

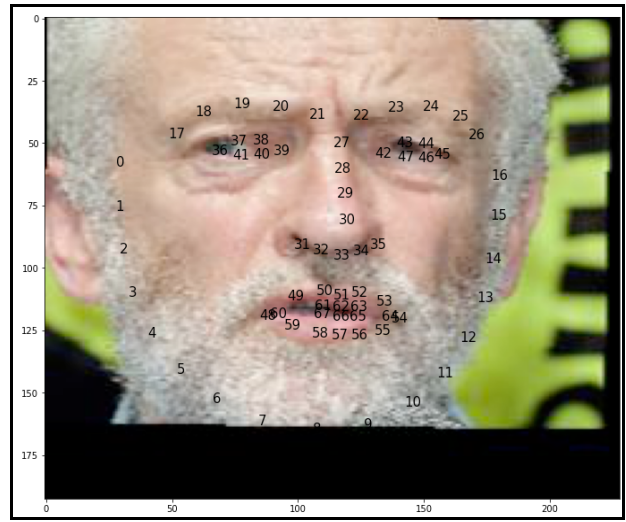

Figure 2. Marked points on face.

These points were then pre-processed by rotating the face to be upright, and scaling all the faces to a comparable size. The machine learning algorithms explored included SVM, XGBoosting, GradientBoosting, and Random Forest.

After resizing all the images to be 200 x 200 pixels, we also moved the landmarks' coordinates accordingly. We also rotated the face to be upright, and measured all distances from the nose. Through trial and error, the set of features used to train the Extreme Gradient Boosting (XGB) model were chosen to be the 2278 normalized pairwise distances between all the landmarks and 29 standardized angles. These include: 9 angles from the outline of the face, 6 angles for the brows, 6 angles for the eyes, 7 angles for the mouth, and 1 angle for the nose.

The XGBoost algorithm produces a good fit on test data, achieving an accuracy of 58.1% on the test set, and an F-1 score of 57.9% (random guessing would be 12.5%). Accuracy levels for individual emotions are as shown in the confusion matrix (see Figure 3). The success rate on "happiness", "disgust", and "contempt" is over 70%. The most difficult emotions for the algorithm were "sadness", "surprise", and fear. The algorithm calculates the probability of each emotion from a face, and then does a forced-choice of the highest-rated emotion as the "answer". Here is the confusion matrix for the results:

Figure 3: Confusion matrix for landmark based approach. True labels on rows; computed labels in columns

| | Neutral | Happiness | Sadness | Surprise | Fear | Disgust | Anger | contempt |
|---|---|---|---|---|---|---|---|---|
| neutral | 0.52 | 0.028 | 0.11 | 0.089 | 0.043 | 0.03 | 0.11 | 0.65 |
| happiness | 0.042 | 0.7 | 0.047 | 0.045 | 0.021 | 0.035 | 0.039 | 0.074 |
| sadness | 0.11 | 0.092 | 0.44 | 0.095 | 0.065 | 0.052 | 0.09 | 0.05 |
| surprise | 0.12 | 0.08 | 0.092 | 0.49 | 0.12 | 0.031 | 0.051 | 0.029 |
| fear | 0.079 | 0.071 | 0.081 | 0.16 | 0.047 | 0.02 | 0.086 | 0.037 |
| disgust | 0.052 | 0.054 | 0.037 | 0.039 | 0.015 | 0.71 | 0.058 | 0.032 |
| anger | 0.12 | 0.047 | 0.086 | 0.033 | 0.047 | 0.067 | 0.54 | 0.063 |
| contempt | 0.047 | 0.046 | 0.024 | 0.042 | 0.01 | 0.0051 | 0.046 | 0.78 |

Figure 4 shows the accuracy plotted against the entropy of the 8 probabilities. Maximum entropy is 2.08. When the entropy is below 0.4 the accuracy is quite good. Sometimes, however, the choices are hard to separate. There are a very few images where the maximum probability is under 0.2 - these images have almost all the emotion probabilities close together, with all of the 8 probabilities being between 0.125 and 0.20.

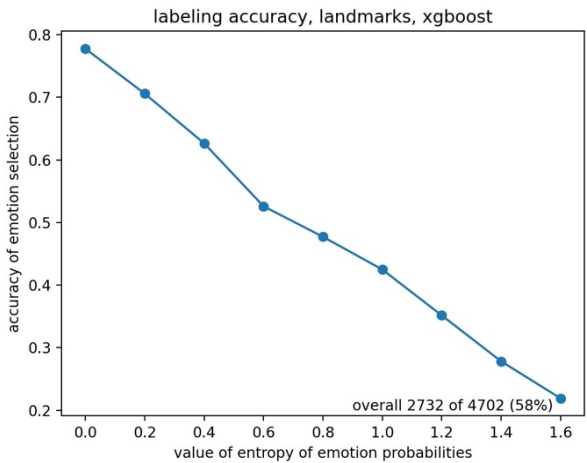

Figure 4. Accuracy vs entropy of prediction.

From the same model, we were able to extract features that were the most significant in identifying emotions. If their contribution to the model was more than the average contribution of all the features then a feature is deemed important. We ended up with 832 important features, 6 of which were angles, and 826 were distances. However, no simple combination of a few features achieves the same overall accuracy.

IV. PIXEL BASED MODELING

Other researchers have recommended deep learning, eg. Coreau et al. (2021) or Ko (2018). Ko, for example suggests CNN methods on single images and LSTM on video sequences. Guo (2018) used the same Affectnet data as this paper, achieving an accuracy of 59% using a CNN method designed to be useful on mobile devices. A large number of methods were compared by Baskar (2017) but without reporting numerical results.

The deep learning model we used was the 3.1M parameter BReG-NeXt-50 model from Hassani et al. (2020) foremotion predictions trained on our training data. The BReG-FNeXt-50 model is a modified ResNet-50 architecture for feature selection with a final fully connected output layer for categorical (or continuous) predictions. This architecture as described is summarized in figure 5 (Figure 2 in the original paper)

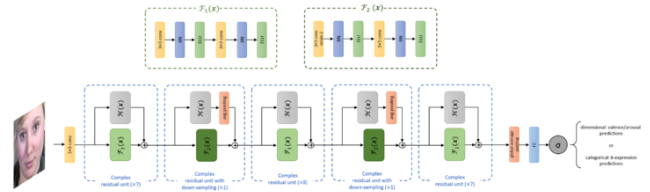

Figure 5. Architecture of deep learning model (Figure 2 in Hassani et al. 2020).

The confusion matrix with deep learning on pixels is shown in Figure 6:

| | neutral | happy | sad | surprise | fear | disgust | anger | contempt |
|---|---|---|---|---|---|---|---|---|
| neutral | 0.55 | 0.02 | 0.13 | 0.10 | 0.03 | 0.02 | 0.08 | 0.06 |
| happy | 0.04 | 0.72 | 0.04 | 0.06 | 0.02 | 0.02 | 0.01 | 0.10 |
| sad | 0.12 | 0.05 | 0.53 | 0.06 | 0.08 | 0.04 | 0.07 | 0.03 |
| surprise | 0.09 | 0.05 | 0.04 | 0.62 | 0.12 | 0.02 | 0.03 | 0.02 |
| fear | 0.04 | 0.03 | 0.08 | 0.23 | 0.52 | 0.05 | 0.03 | 0.01 |
| disgust | 0.04 | 0.06 | 0.12 | 0.03 | 0.03 | 0.55 | 0.12 | 0.05 |
| anger | 0.09 | 0.03 | 0.13 | 0.03 | 0.05 | 0.10 | 0.55 | 0.04 |
| contempt | 0.15 | 0.10 | 0.05 | 0.03 | 0.00 | 0.02 | 0.01 | 0.56 |

Figure 6. Confusion matrix with deep learning.

Figure 7 displays correctness against the entropy of the choices. Again, we can see how the reliability of an emotion assignment is shown by the apparent confidence of the numerical results.

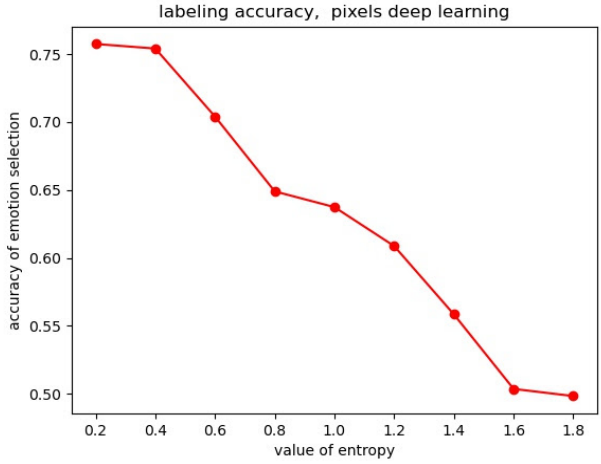

Figure 7. Accuracy vs. entropy of prediction.

## V. A META-CLASSIFIER

Both methods work well on the same 4520 test images, 2636 are correct as classified by landmarks, and 2611 are correct as classified by pixels. Both methods agree on 1726 of the images (whether right or wrong). When the methods disagree (2794 faces), the higher prediction value is right 1949 times, the lower predicted value is right 314 times, and neither method is right on 531 faces. Thus, these two methods, in spite of their similar overall accuracy, perform differently in different images as indicated by the table below based on the test data. In Figure 8, the rows are classified by the DL method while the columns are the classification chosen by the XGBoost algorithm. The cell entry *(i,j)* tells: for the DL classification of *i*th type of image, how many were classified by *j*th image classification using XGBoost. The differences between them are striking. Cohen's *kappa*, which measures the agreement between these two models is only 30.14%, indicating only fair agreement (McHugh 2012).

|  | neutral | happy | sad | surprise | fear | disgust | anger | contempt |
|---|---|---|---|---|---|---|---|---|
| neutral | 0.30 | 0.034 | 0.15 | 0.10 | 0.062 | 0.059 | 0.13 | 0.16 |
| happy | 0.04 | 0.57 | 0.054 | 0.051 | 0.029 | 0.072 | 0.048 | 0.13 |
| sad | 0.16 | 0.085 | 0.23 | 0.092 | 0.084 | 0.12 | 0.15 | 0.076 |
| surprise | 0.12 | 0.10 | 0.11 | 0.32 | 0.18 | 0.047 | 0.055 | 0.057 |
| fear | 0.11 | 0.076 | 0.125 | 0.21 | 0.29 | 0.04 | 0.11 | 0.03 |
| disgust | 0.066 | 0.075 | 0.059 | 0.047 | 0.061 | 0.054 | 0.094 | 0.045 |
| anger | 0.14 | 0.04 | 0.092 | 0.048 | 0.059 | 0.14 | 0.35 | 0.13 |
| contempt | 0.07 | 0.11 | 0.048 | 0.042 | 0.023 | 0.044 | 0.065 | 0.57 |

Figure 8. Results of image categorization with deep learning (Rows) and XGBoost (columns)

Given these striking differences, we experimented with creating a *meta-classifier* by combining the two methods. We considered seeing if one method was generally better for one emotion rather than another, but the best method, which achieved a 76.70% accuracy and 76.83% F-1 score, weighted based the apparent confidence of each method. The technique is to add the probability outputs from each of the two methods and re-normalize so that the combined probabilities sum to 1.

That is, for a given picture suppose BReG-NeXt and XGBoost models outputs are $p_i^{DL}$ and $p_i^{XGB}$, $i=1,...,8$, respectively, as predicted probabilities for each of the 8 emotion categories. Then the combined predicted probability for each category would then be

$$q_i = \frac{\exp(p_i^{DL} + p_i^{XGB})}{\sum_{j=1}^{8} \exp(p_j^{DL} + p_j^{XGB})}$$

for each emotion category corresponding to ['neutral', 'happy', 'sad', 'surprise','fear', 'disgust', 'anger', 'contempt'] in that order. The forced-choice category would then be *arg max*{$q_i$}. Choosing the lowest-entropy choice for each image is almost as effective.

The accuracy of this *meta-classifier* is .0.767 with the F1-score of .0.768. The test confusion matrix combining the landmark and pixel algorithms is shown in Table 9.

|  | neutral | happy | sad | surprise | fear | disgust | anger | contempt |
|---|---|---|---|---|---|---|---|---|
| neutral | 0.74 | 0.08 | 0.074 | 0.067 | 0.009 | 0.003 | 0.067 | 0.034 |
| happy | 0.020 | 0.84 | 0.018 | 0.033 | 0.006 | 0.015 | 0.008 | 0.059 |
| sad | 0.082 | 0.048 | 0.69 | 0.053 | 0.046 | 0.026 | 0.043 | 0.019 |
| surprise | 0.067 | 0.059 | 0.027 | 0.73 | 0.075 | 0.016 | 0.023 | 0.007 |
| fear | 0.026 | 0.031 | 0.052 | 0.16 | 0.68 | 0.021 | 0.035 | 0.002 |
| disgust | 0.025 | 0.023 | 0.029 | 0.008 | 0.014 | 0.85 | 0.035 | 0.020 |
| anger | 0.058 | 0.018 | 0.070 | 0.012 | 0.021 | 0.060 | 0.74 | 0.016 |
| contempt | 0.036 | 0.026 | 0.009 | 0.006 | 0.00 | 0.002 | 0.025 | 0.90 |

Table 9. Confusion matrix for the combined algorithm.

## VI. CONCLUSION AND SUMMARY

This paper experimented with prediction of 8 categories of emotions extracted from a balanced subset of manually annotated faces from the well-known Affectnet database. Two supervised learning methods were used: namely a deep-learning resnet model based on all pixels, and XG-boost model based on 68 landmarks on faces. Both produced comparable results (around 58% accuracy on test data). However, a new meta-classifier, combining the two methods produced remarkably better results (77% accuracy on test data). The results suggest that the meta-classifier approach is likely to perform better than either of deep-learning or statistical-learning methods.

ACKNOWLEDGMENT

We appreciate the assistance of Vikki Sui.

REFERENCES

[1] Ekman, P. (1993). Facial Expression and Emotion. American Psychologist, 48 (4), 384-392.

[2] W. Mellouk, W. Handouzi, Facial emotion recognition using deep learning: review and insights, Procedia Computer Science 175 (2020) 689-694.

[3] Harms, M.B., Martin, A. & Wallace, G.L. Facial Emotion Recognition in Autism Spectrum Disorders: A Review of Behavioral and Neuroimaging Studies. *Neuropsychol Rev* **20,** 290–322 (2010).

[4] M. Jeong and B. C. Ko, "Driver's Facial Expression Recognition in Real-Time for Safe Driving," *Sensors*, vol. 18, no. 12, p. 4270, Dec. 2018.

[5] P. Naga, Swamy Das Marri and R. Borreo, Facial emotion recognition methods, datasets and technologies: A literature survey, Materials Today: Proceedings, https://doi.org/10.1016/j.matpr.2021.07.046

[6] Heng-Tze Cheng et al (2016), "Wide & Deep Learning for Recommender Systems," *DLRS 2016: Proceedings of the 1st Workshop on Deep Learning for Recommender Systems*, Sep 2016, pp 7-10.

[7] Jianxun Lian, Xiaohuan Zhou, Fuzheng Zhang, Zhongxia Chen, Xing Xie, and Guangzhong Sun. 2018. XDeepFM: Combining Explicit and Implicit Feature Interactions for Recommender Systems. In .*Proceedings of the 24th ACM SIGKDD International Conference on Knowledge Discovery & Data Mining* (*KDD '18*). Association for Computing Machinery, New York, NY, USA, 1754–1763.

[8] B. Hassani, P. S. Negi, and M. H. Mahoor (2020). "Breg-next: Facial affect computing using adaptive residual networks with bounded gradient." https://arxiv.org/abs/2004.08495.

[9] Zhu X, Liu X, Lei Z, Li SZ. Face Alignment in Full Pose Range: A 3D Total Solution. IEEE Trans Pattern Anal Mach Intell. 2019 Jan;41(1):78-92

[10] Su, Yanchao & Ai, Haizhou & Lao, Shihong. (2009). Multi-View Face Alignment Using 3D Shape Model for View Estimation. 179-188. Advances in Biometrics, Third International Conference, ICB 2009, Alghero, Italy, June 2-5.

[11] Rachel Coreau, Ian Pépin, Cameron Duffy, Hazem M. Abbas, and Hossam Hassanein. 2021. Deep learning models and techniques for facial emotion recognition. In *Proceedings of the 31st Annual International Conference on Computer Science and Software Engineering* (*CASCON '21*). IBM Corp., USA, 14–22.

[12] Ko, B. C. A Brief Review of Facial Emotion Recognition Based on Visual Information. *Sensors* (Basel). 2018 Jan 30;18(2):401.

[13] Guo, Yuanyuan et al. "Real-Time Facial Affective Computing on Mobile Devices." *Sensors (Basel, Switzerland)* vol. 20,3 870. 6 Feb. 2020,

[14] Baskar A., Gireesh Kumar T. (2018) Facial Expression Classification Using Machine Learning Approach: A Review. In: Satapathy S., Bhateja V., Raju K., Janakiramaiah B. (eds) Data Engineering and Intelligent Computing. Advances in Intelligent Systems and Computing, vol 542. Springer, Singapore.